\pdfoutput=1
\documentclass[11pt]{article}

\usepackage{acl}

\usepackage{times}
\usepackage{latexsym}

\usepackage[T1]{fontenc}
\usepackage[utf8]{inputenc}

\usepackage{microtype}
\usepackage{amsmath}
\usepackage{amssymb}
\usepackage{amsfonts}
\usepackage{graphicx}
\usepackage{subfigure} 
\usepackage{booktabs}
\usepackage{multirow}
\usepackage{svg}
\usepackage{algorithm}
\usepackage{algpseudocode}
\usepackage{amsmath}
\usepackage{algpseudocode}
\usepackage{algorithm}
\usepackage{subfigure}
\usepackage{caption}
\algnewcommand\algorithmicforeach{\textbf{for each}}
\algdef{S}[FOR]{ForEach}[1]{\algorithmicforeach\ #1\ \algorithmicdo}
\definecolor{grassgreen}{rgb}{0.5, 1.0, 0.0}

\title{Reasoning before Responding: Integrating Commonsense-based Causality Explanation for Empathetic Response Generation}

\author{Yahui Fu, Koji Inoue, Chenhui Chu, and Tatsuya Kawahara \\ Graduate School of Informatics, Kyoto University, Japan\\ \texttt{[fu, inoue, kawahara]@sap.ist.i.kyoto-u.ac.jp}\\
\texttt{chu@i.kyoto-u.ac.jp}}
\begin{document}
\maketitle
\begin{abstract}
% Empathy plays a crucial role in enabling machines to comprehend and respond appropriately to users' emotions and experiences in human-machine conversations. 
Recent approaches to empathetic response generation try to incorporate commonsense knowledge or reasoning about the causes of emotions to better understand the user's experiences and feelings. However, these approaches mainly focus on understanding the causalities of context from the user's perspective, ignoring the system's perspective. In this paper, we propose a commonsense-based causality explanation approach for diverse empathetic response generation that considers both the user's perspective (user's desires and reactions) and the system's perspective (system's intentions and reactions). We enhance ChatGPT's ability to reason for the system's perspective by integrating in-context learning with commonsense knowledge. Then, we integrate the commonsense-based causality explanation with both ChatGPT and a T5-based model. Experimental evaluations demonstrate that our method outperforms other comparable methods on both automatic and human evaluations.

\end{abstract}

\section{Introduction}
Empathy is a desirable capacity of humans to place themselves in another’s position
to show understanding of his/her experience and feelings and respond appropriately. Empathy involves both cognitive and affective aspects \cite{davis1983measuring}, including the ability to perceive the user's situation and express appropriate emotions.

Previous work on empathetic response generation has primarily focused on the affective aspect of emotional expression \cite{lin2019moel,majumder2020mime,li2020empdg} by emotion detection, without sufficient consideration of context understanding. Recently, there has been a growing interest in exploring context understanding by leveraging external commonsense knowledge for reasoning emotion causes-effects or the user's desires, such as \citet{sabour2022cem} and \citet{wang2022empathetic,wang2022care}. 
\begin{figure}[H]
  \centering
  \subfigure[Example of using commonsense from COMET to generate a response from the user's perspective.]{
    \includegraphics[width=0.45\textwidth]{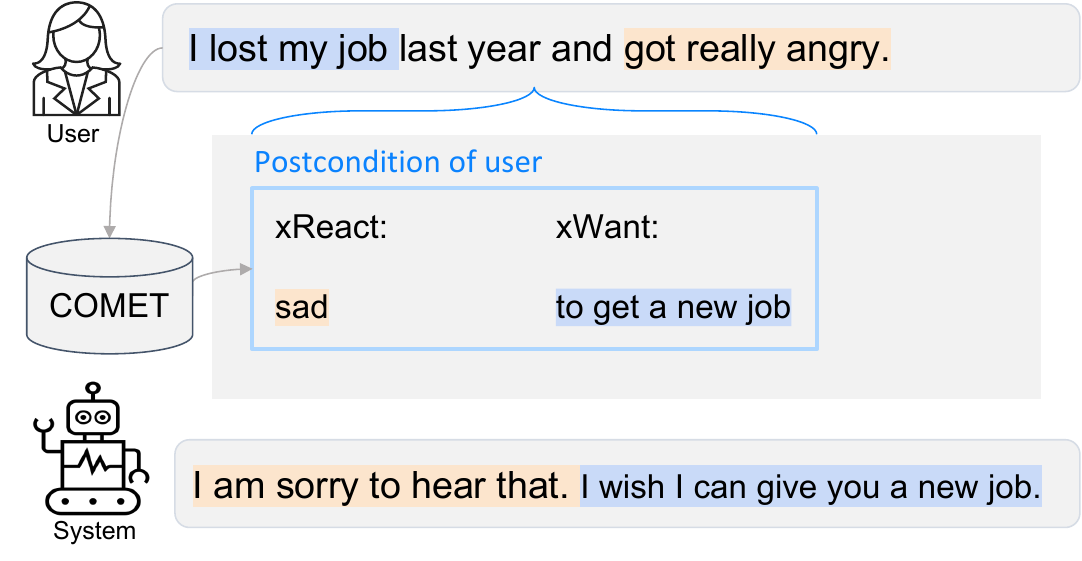}
    \label{fig:subfigure1}}
  
  \vspace{0.01cm} % Adjust the vertical spacing between subfigures
  
  \subfigure[Example of a response from the actual responder's perspective, based on reasoning reaction and intent to mimic humans.
  % where the actual responder's intention is not aligned with the user's desire.
  ]{
    \includegraphics[width=0.45\textwidth]{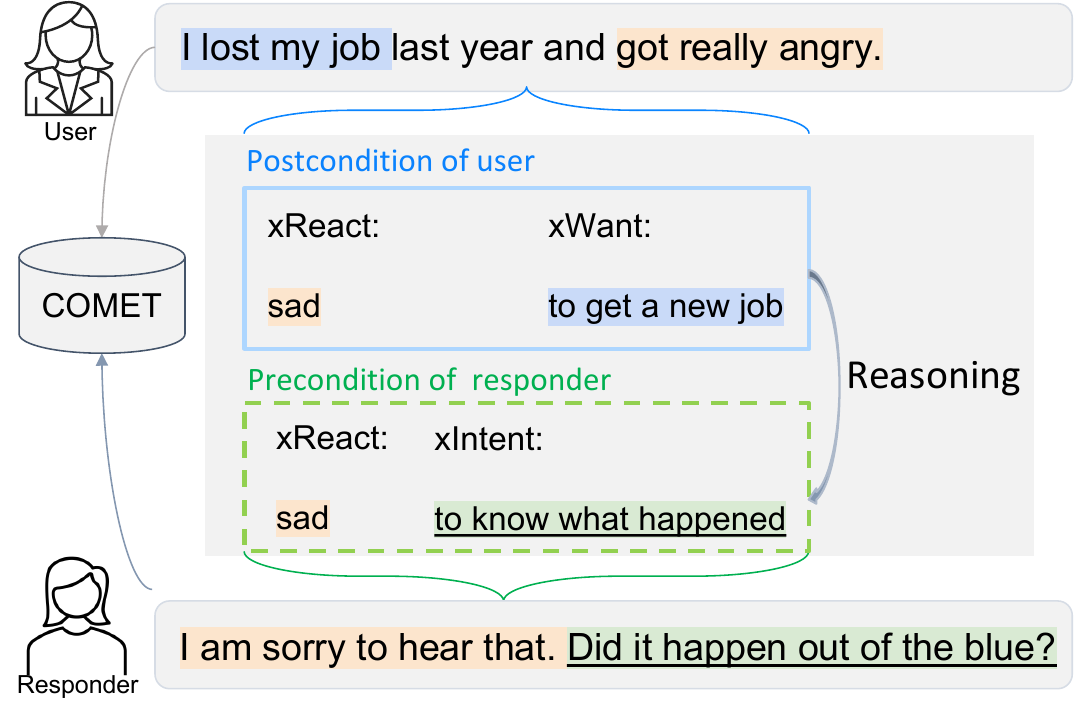}
    \label{fig:subfigure2}}
  \caption{Two examples to produce a response from different perspectives. The blue solid box contains "xReact" and "xWant" representing the user's emotional reaction and desires. The green dotted box comprises "xReact" and  "xIntent," representing the emotional reaction and intention of the actual responder.}
  \label{fig:mainfigure}
\end{figure}
\noindent However, these approaches focus on understanding the causalities from the user's perspective. %neglecting the system's perspective. As depicted in Figure \ref{fig:my_label1}, Knowing the user's desires can directly lead to a response that satisfies the user. 

Exploring the causality within the user's context and reasoning his/her desires can be helpful so that the system's intention is aligned with the user's desires, and the response is generated from the user's perspective (Figure \ref{fig:subfigure1}).
However, in real human communication, the responder's intention is not always confined to the user's desires, as shown in Figure \ref{fig:subfigure2}. Relying solely on the user's desire to generate a response may not fully understand the user's experience, and leads to weak empathy, as shown in Figure \ref{fig:subfigure1}.
Therefore, it is necessary to incorporate both the user's perspective (exploring his/her desire and reaction) and the system's perspective (reasoning its intention and reaction to mimic humans) for empathetic response generation.%\citet{welivita2020taxonomy} introduced the eight most common empathetic listener intents, including questioning, acknowledging, agreeing, consoling, encouraging, sympathizing, wishing, and suggesting. The intention of the system's utterance, as shown in Figure \ref{fig:my_label1}, can be classified as consoling or sympathizing, while the intent of sys's utterance, as shown in Figure \ref{fig:my_label2} can be classified as questioning. 
% Therefore, to generate more diverse empathetic responses, it is important to explore not only the user's desires but also the system's intentions, as well as the emotions. 

Through the utilization of COMET \cite{bosselut2019comet}, which is a pre-trained GPT-2 model (Radford et al. 2018) fine-tuned on the if-then reasoning graph from ATOMIC \cite{sap2019atomic}, the system's possible intentions can be predicted to align with the user's desires. However, the system's intention may not be constrained by the user's desire. Therefore, we do not adopt COMET for the system's intention reasoning.

ChatGPT\footnote{\url{https://chat.openai.com/}} has shown its efficacy in several tasks \cite{zhao2023chatgpt}. \citet{bang2023multitask} introduced ChatGPT's potential in causal reasoning on human-annotated explainable CAusal REasoning dataset (E-CARE) \cite{du2022care}. However, it is based on whether the model can make a judgment on correct causes or effects instead of generating causality explanations. In this paper, we propose to enhance it by incorporating in-context learning with commonsense reasoning for causality explanation. Our main contributions are as follows:
\begin{itemize}
    \item We propose to integrate a commonsense-based causality reasoning for empathetic response generation, which takes the system's intention and reaction, along with the user's desire and reaction.
    \item We propose to enhance ChatGPT's capability for causality explanation through the integration of in-context learning with commonsense knowledge (desire, reaction, and intention). 
    \item We present experimental results to demonstrate both ChatGPT and a T5-based model, integrated with the proposed commonsense-based causality explanation, outperform other competitive methods based on both automatic and human evaluations.
\end{itemize}

\section{Related Work}
\subsection{Commonsense and Causality Reasoning for Empathetic Response Generation}
\citet{kim2021perspective} extracted emotion causes from the dialogue
context by utilizing a rational speech act framework. \citet{sabour2022cem,wang2022empathetic} utilized ATOMIC-2020 \cite{hwang2021comet}, which is a collection of commonsense reasoning inferences about everyday if-then events, to enrich context understanding with information on the user's reactions, intentions, effects, needs, and desires. %Additionally, \citet{wang2022care} employs a cause-effect graph \cite{li2021guided} to reason about the emotion causes and effects, thereby improving context understanding. 
However, these approaches only focus on understanding the causalities within the context from the user's perspective for empathetic response generation, ignoring the system's perspective. 
% However, these approaches for empathetic response generation are only from the user's perspective, which does not consider the system's perspective. 

\subsection{Large Language Models for Empathetic Response Generation}
With the development of large language models such as GPT-3 \cite{brown2020language} and ChatGPT, many studies have shown their ability on various NLP tasks with either a few-shot or zero-shot setting \cite{madotto2021few,lee2022does,zhao2023chatgpt}. 
\citet{lee2022does} introduced two selection methods that choose in-context examples
based on emotion and situation information to generate empathetic responses by GPT-3. 
\citet{zhao2023chatgpt} showed ChatGPT's ability on empathetic response generation.
% and showed that ChatGPT outperforms the SOTA EmpSOA \cite{zhao2022don} model on the aspects of \textit{Coherence}, \textit{Empathy} and \textit{Informativeness} by human evaluation. 
In this study, we enhance ChatGPT with a commonsense-based causality explanation prompt for empathetic response generation.

% emotion-cause recognition. They evaluate the performance of ChatGPT and all baseline models on the RECCON-DD dataset \cite{poria2021recognizing}, which features utterance-level emotion labels and emotion cause labels  based on
% the DailyDialog dataset \cite{li2017dailydialog}. and they report that ChatGPT 
%  holds an 11.95\% gap in terms of macro F1 score compared to the SOTA-supervised
% method.

\section{Preliminaries}

\subsection{Knowledge Acquisition}
% The information contained within the blue and green boxes represents relevant knowledge inferred from the utterances of the user and system, respectively, using a modified BART-based \cite{lewis2019bart} variation of COMET \cite{bosselut2019comet}, which is trained on the ATOMIC-2020 dataset \cite{hwang2021comet}. 

In order to  generate commonsense inferences for given events, we adopt a modified BART-based \cite{lewis2019bart} variation of COMET, which was trained on the ATOMIC-2020 dataset \cite{hwang2021comet}. This model is suitable for inferring knowledge regarding unseen events \cite{hwang2021comet}, like events in the EmpatheticDialogue dataset \cite{rashkin2018towards}. 

% We leverage this model to infer three commonsense relations for the person involved in the event: their reaction to the event (xReact), their intent before the event (xIntent), and what they would desire after the event (xWant). 
In the training process, we leverage this model to infer the relations of \textit{xWant} and \textit{xReact} for each user's utterance in the training set and the relations of \textit{xIntent} and \textit{xReact} for the system's utterance, which are inferred from the ground-truth response in training. In the testing, we only infer the relations of \textit{xWant} and \textit{xReact} for the user's utterance. The system's \textit{xIntent} and \textit{xReact} will be inferred by the proposed causality reasoning module.

\begin{figure*}
  \centering

  \subfigure[Proposed causality reasoning module and enhanced ChatGPT-based empathetic response generation method.]{
    \includegraphics[width=\textwidth]{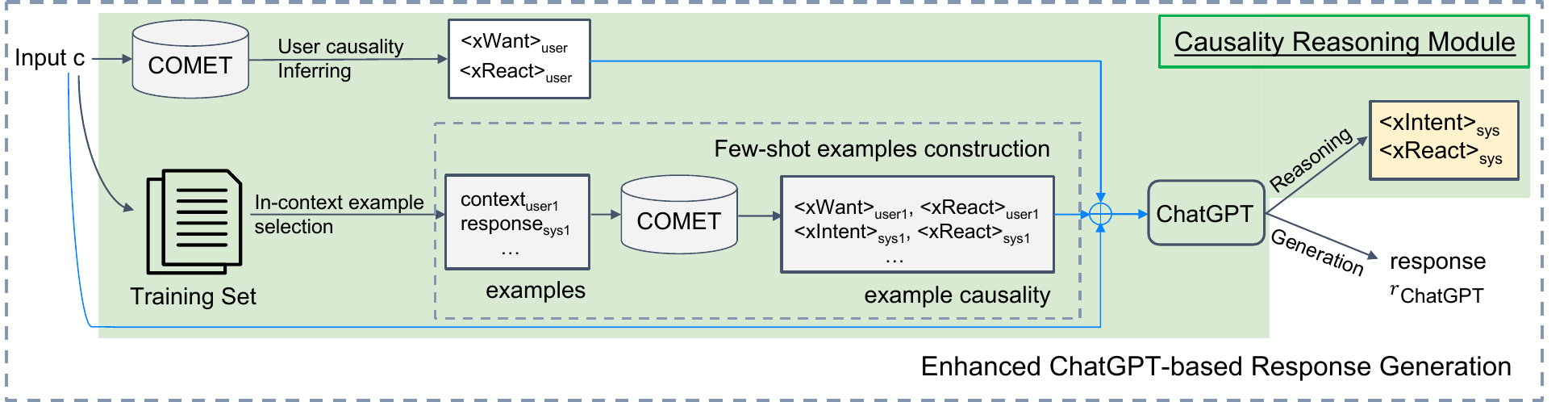}
    \label{fig:subfigure3}}
  
  \vspace{0.01cm} % Adjust the vertical spacing between subfigures
  
  \subfigure[Integrating the causality reasoning module into a T5-based encoder-decoder for empathetic response generation.]{
    \includegraphics[width=0.99\textwidth]{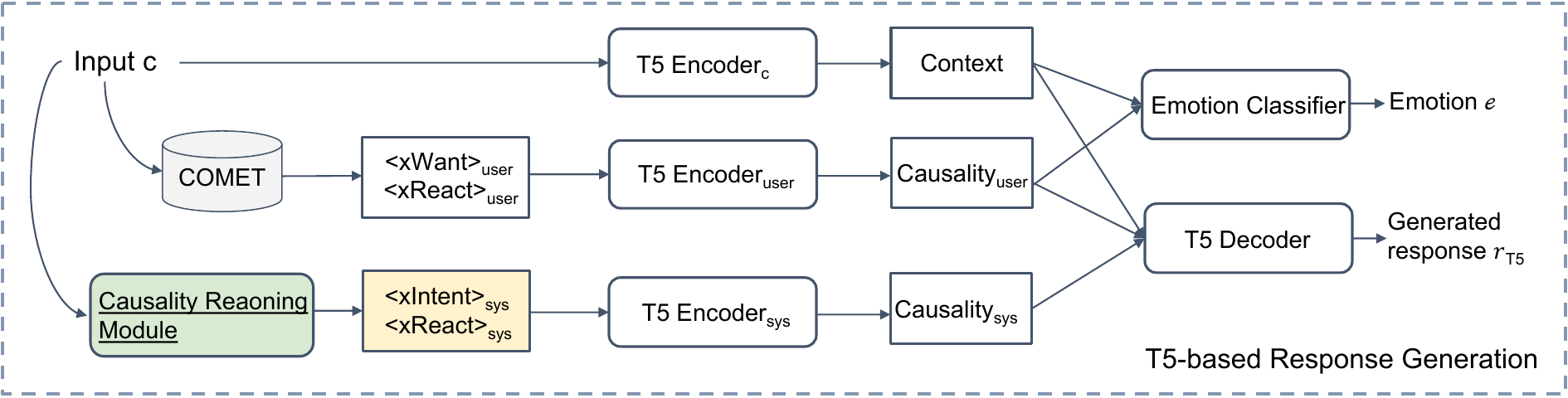}
    \label{fig:subfigure4}}

  \caption{Overview of our proposed model. The input $c$ ends with the user's utterance. The generated response $r_{T5}$ and $r_{ChatGPT}$ are in the role of the system (sys).}
  \label{fig:mainfigure}
\end{figure*}

\subsection{In-Context Example Selection} \label{sec3.2}
% The performance of GPT3 or ChatGPT is sensitive to the quality of in-context examples. 
We enhance ChatGPT's causality explanation based on the few-shot setting. Given the sensitivity of large language models such as ChatGPT to in-context examples \cite{liu2021makes,lee2022does}, we adopt a method similar to \citet{lee2022does} to select top-$k$ examples from the training set based on the similarity between the test conversation and the training conversations. %The situation utterance describes the topic and summary of each conversation.
Specifically, we adopt Sentence BERT introduced by \citet{reimers-2019-sentence-bert} to encode the sentence semantics of the conversation. In this study, we compute the cosine similarity between the situation utterance of the training set and the test sample, which is annotated in the dataset.
Top-$k$ samples are chosen from the training set for each test sample as in-context few shot examples for ChatGPT. %$k\in n$. 
% for each $(s_{i},c_{i})$ in the test set, and $(s_{n},c_{n},r_{n})$ in the training set, we compute the cosine similarity between $s_{i}$ and $s_{n}$. Here, $s$, $c$, and $r$ represent conversation situation, input utterance, and ground truth response, respectively.
% \begin{equation}
%     sim_{i,n}=\frac{SBERT(s_{i})\cdot SERT(s_{n})}{\left\| SBERT(s_{i}) \right\|_{2}\left\| SBERT(s_{n}) \right\|_{2}}
% \end{equation}

\section{Proposed Method}
Figure \ref{fig:mainfigure} shows an overview of our proposed method. It consists of three components: (1) Causality reasoning module, which aims to enhance the ChatGPT or T5 decoder with a causality explanation for empathetic response generation. %Table \ref{tab:my-table} further explains the few-shot examples construction in this module.
(2) Enhanced ChatGPT-based response generation. (3) T5-based response generation, which is based on a trained T5 encoder-decoder to be compared with other approaches that have developed their own model using the EmpatheticDialogue dataset \cite{lin2019moel,majumder2020mime,li2020empdg,sabour2022cem,majumder2022exemplars}.
% which contains two key parts, one is the T5 encoder-decoder, and another is the causality learning module to enrich the T5 decoder with causality explanation. We will explain each module in the following.

\subsection{Causality Reasoning Module based on ChatGPT} \label{sec:learning}
As outlined in Algorithm \ref{alg:cap}, this module consists of four steps. Initially, for a test input $c$, we employ the method outlined in Section \ref{sec3.2} to select the top-$k$ relevant training samples, denoted as $\mathcal S$, for in-context learning, 
such as (context1, response1) and (context2, response2) as exemplified in Table \ref{tab:my-table18} in Appendix \ref{sec:introduction}.

In the second step, for each selected sample $(c_{n},r_{n})\in \mathcal S$, we leverage the COMET model to infer the \textit{xWant} ($c_{nWant}$) and \textit{xReact} ($c_{nReact}$) knowledge corresponding to the user's utterance $c_{n}$. Additionally, we extract the \textit{xIntent} ($r_{nIntent}$) and \textit{xReact} ($r_{nReact}$) knowledge pertaining to the ground truth system response $r_{n}$. This information is then concatenated as few-shot examples (Table \ref{tab:my-table18} in Appendix \ref{sec:introduction}), denoted as $\mathcal{M}_{prompt}$.

Thirdly, for the test input $c$, we obtain the \textit{xWant} ($c_{Want}$) and \textit{xReact} ($c_{React}$) knowledge using COMET. Finally, they are appended to $\mathcal{M}_{prompt}$ as the prompt to ChatGPT, which reasons \textit{Intent} ($r_{Intent}$) and \textit{React} ($r_{React}$) from the system's perspective based on the few-shot learning.
% taking into account the input context of the test sample ($c$).
% we construct $\mathcal{M}{prompt}^+$, which represents the prompt template utilized by ChatGPT. 

\subsection{Enhanced ChatGPT-based Response Generation}
The prompt provided to ChatGPT encompasses two components: causality explanation from the user's perspective, predicted by COMET, and causality explanation from the system's perspective, derived through the causality reasoning module described in Section \ref{sec:learning}. These components, along with the few-shot examples, are integrated into ChatGPT to generate empathetic responses.

\begin{algorithm}
\caption{Commonsense-based causality explanation prompt}
\label{alg:cap}
\begin{algorithmic}
\Require A training set $\mathcal{D}$=\{(c$_{n}$,r$_{n}$)\}$_{n=1}^N$, $N$ is the number of training samples; a test input ($c$); $c$, $r$ represents context, ground truth response, respectively; COMET model $f_{\theta}\left( \cdot  \right)$
% \Ensure $y = x^n$
\State \textbf{ {/*Step 1: In-context examples selection*/}}
\State $\mathcal{M}_{sim} \gets$ empty list
\ForEach{$d$=(c$_{n}$,r$_{n}$) $\in \mathcal D$}
\State Get similarity score: $sim_{n}$
\State $\mathcal{M}_{sim}$.append($sim_{n}$)
\EndFor
\State $\mathcal{S}$=\{(c$_{n}$,r$_{n}$)\}$_{n=1}^k$=max($\mathcal{M}_{sim}$,$k$), $k$ is the number of in-context examples
\State \textbf{ {/*Step 2: Get the commonsense knowledge for the selected examples */}}
\State $\mathcal{M}_{prompt} \gets$ empty list
\ForEach{$s$ $\in \mathcal{S}$}
\State Get causality information (desire and reaction of user, intent, and reaction of sys) for the sample in $\mathcal{S}$ inferred by COMET
\State $c_{nWant}$= $f_{\theta}\left(c_{n}+[xWant]\right)$
\State $c_{nReact}$=$f_{\theta}\left(c_{n}+[xReact]\right)$
\State $r_{nItent}$=$f_{\theta}\left(r_{n}+[xIntet]\right)$
\State $r_{nReact}$=$f_{\theta}\left(r_{n}+[xReact]\right)$
\State $k_{n}$=$c_{nWant}$+$c_{nReact}$+$r_{nIntent}$+$r_{nReact}$
\State $\mathcal{M}_{prompt}$.append($c_{n}$,$k_{n}$,$r_{n}$)
\EndFor
\State \textbf{{/*Step 3: Get the commonsense knowledge for the test sample */}}
\State Get causality information (desire and reaction of user) for the test sample $c$
\State $c_{Want}$= $f_{\theta}\left(c+[xWant]\right)$
\State $c_{React}$=$f_{\theta}\left(c+[xReact]\right)$
\State \textbf{ {/*Step 4: prompting ChatGPT, and output the reasoned \textit{Intent}, \textit{React} for generating a empathetic response*/}}
\State Input: $\mathcal{M}_{prompt}^+$=$\mathcal{M}_{prompt}$+$c$+$c_{Want}$+$c_{React}$
\State Output: $r_{Itent}$, $r_{React}$, $r_{ChatGPT}$
\end{algorithmic}
\end{algorithm}

\subsection{T5-Based Response Generation}
\noindent\textbf{Context and Causality Encoding} For a test input $c$, we use the COMET model to infer the user's causality information, which are desire and reaction of the user ($k_{user}$: $c_{Want}$ and $c_{React}$), and use the causality reasoning module based on ChatGPT to infer the system's causality information, which are intention and reaction of the system ($k_{sys}$: $r_{Itent}$, $r_{React}$).
We utilize three T5 encoders for encoding input context, the user's causality information, and the system's causality information. 
\begin{equation}
\begin{aligned}
z_{c}&=T5_{enc}^{c}(c)\\
z_{user}&=T5_{enc}^{user}(k_{user})\\
z_{sys}&=T5_{enc}^{sys}(k_{sys})
\end{aligned}
\end{equation}

\noindent\textbf{Emotion Classification}
In order to detect the user’s affective state, we concatenate the context representations and the user's causality information, and then pass them through a linear
layer followed by a softmax operation to produce the emotion category distribution:
\begin{equation}
    p_{e} = softmax(W_{e}(z_{c}\oplus z_{user})) 
\end{equation}
where $W_{e}$ is the weight vector of the linear layer. Given the ground-truth emotion label $e^*$ for each conversation, the cross-entropy loss is computed to optimize the process of emotion classification:
\begin{equation}
    \mathcal{L}{e}= -\log (p_{e}(e^*))
\end{equation}

\noindent\textbf{Response Generation}
We fuse and feed the information of the user's context and the corresponding causality explanation of the user and the system to a fully-connected (FC) layer.
\begin{equation}
    z_{fused}=FC([z_{c} \oplus z_{user} \oplus z_{sys}])
\end{equation}
Subsequently, the target response $r_{T5}$ = [$y_{1}$,...,$y_{T}$ ] with length $T$, is generated by the T5 decoder token by token:
\begin{equation}
        p\left( y_{t}|c, y_{<t}\right)=T5^{c}_{dec}(E_{y<t},z_{fused})
\end{equation}
where $E_{y<t}$ denotes the embeddings of the tokens that have been generated.
The negative log-likelihood for generation is defined as:
\begin{equation}
    \mathcal{L}_{gen}=-\sum_{t=1}^{T}\log p\left( y_{t}|c, y_{<t}\right)
\end{equation}
The combined loss is defined as:
\begin{equation}
    \mathcal{L}=\mathcal{L}_{e}+\mathcal{L}_{gen}
\end{equation}
% $\gamma_{1}$ and $\gamma_{2}$ are the hyper-parameters that we use to
% control the influence of each loss; in our experiments, we set them to 1.

\begin{table*}[t]
\caption{Evaluations of reaction and intention reasoned by ChatGPT+Causality$_{user,sys}$, and we set the corresponding knowledge of ground-truth response inferred by COMET as the reference. PBert, RBERT, and FBert represent Bertscore in terms of precision, recall, and F1, respectively.}
\scalebox{0.75}{
\begin{tabular}{@{}lrrrrrrrrrrrrrr@{}}
\toprule
\multirow{2}{*}{} & \multicolumn{7}{c}{Reaction}                                                   & \multicolumn{7}{c}{Intention}                              \\ \cmidrule(l){2-15} 
k          & F1    & BLEU-2 & BLEU-3 & BLEU-4 & PBert & RBert & \multicolumn{1}{r|}{FBert} & F1    & BLEU-2 & BLEU-3 & BLEU-4 & PBert & RBert & FBert \\ \midrule
2               & 19.32 & 6.81   & 3.16   & 1.56   & 91.92 & 92.60 & \multicolumn{1}{r|}{92.25}  & 13.29 & 14.65  & 6.39   & 3.49   & 88.90 & 89.17 & 89.02 \\
3               & 21.83 & 7.12   & 3.25   & 1.34   & 92.28 & 92.74 & \multicolumn{1}{r|}{92.50}  & 14.49 & 17.39  & 8.91   & 5.37   & 89.13 & 89.40 & 89.26 \\
4               & 25.83 & 8.74   & 3.72   & 1.48   & 92.55 & 92.92 & \multicolumn{1}{r|}{92.73}  & 15.14 & 19.05  & 10.07  & 6.14   & 89.30 & 89.54 & 89.41 \\
5               & 27.87 & 8.52   & 3.55   & \textbf{1.69}   & 92.76 & 92.95 & \multicolumn{1}{r|}{92.85}  & 15.00 & 19.74  & 10.69  & 6.51   & 89.29 & 89.46 & 89.37 \\
6               & \textbf{29.53} & \textbf{9.43}   & \textbf{4.14}   & 0.00   & \textbf{93.15} & \textbf{93.22} & \multicolumn{1}{r|}{\textbf{93.18}}  & \textbf{15.71} & \textbf{20.72}  & \textbf{11.55}  & \textbf{7.25}   & \textbf{89.62} & \textbf{89.76} & \textbf{89.68} \\ \bottomrule
\end{tabular}}
\label{tab:my-table1}
\end{table*}

\section{Evaluation of Causality Explanation based on ChatGPT}
We first evaluate how the output of the causality reasoning module is matched with the reaction and intention of the actual (ground-truth) response.
\subsection{Dataset}
The EmpatheticDialogues dataset of 25k empathetic conversations is used.
% between a speaker and a listener grounded in an emotional situation. 
% The dataset provides 32 evenly distributed emotion labels which are common in daily chats. 
The ratio for training/validation/test is 8:1:1.

\subsection{Setting}
% For the experiments of GPT-3 and GPT-3+Causality$_{user,sys}$, we utilize the "davinci" engine version with a temperature of 1, top p of 0.5, and frequency and presence penalties set to 0. 
For the experiments based on ChatGPT, we used the "gpt-3.5-turbo" engine version with a temperature of 0. We used the 10\% of the EmpatheticDialogue test set for this evaluation (250 samples for single-turn and multi-turn settings, respectively).

\subsection{Automatic Metrics} \label{T5 automatic}
\noindent\textbf{(Macro-averaged) F1 score} \cite{rajpurkar2016squad}, precision, and recall are computed by matching the portion of words in the generation and ground truth that overlap after removing stopwords. 

\noindent\textbf{BLEU} \cite{papineni2002bleu} evaluates the matching between n-grams of the generated response to the ground truth. We utilize BLEU-2, BLEU-3, and BLEU-4 scores. 

\noindent\textbf{BERTScore } \cite{zhang2019bertscore} is a BERT-based evaluation measure for text generation, which focuses on lexical semantic similarity between the generated response and the ground truth. We adopt its precision, recall, and F1 score (PBERT, RBERT, FBERT). We used the RoBERTa-Large \cite{liu2019roberta} version.

\subsection{Results}
We evaluate the performance of the system's intention/reaction reasoning under a different number of in-context examples. Experimental results in Table \ref{tab:my-table1} show that increasing the value of $k$ allows for ChatGPT to generate reactions and intentions that are more closely aligned with those inferred by COMET from the ground truth response.
% but still a big gap to the ground truth if when $k$ set to 6.

% \begin{table}[ht]
% \caption{Ablation study on the number of in-context examples $k$ in the prompt.}
% \begin{tabular}{@{}lccccc@{}}
% \toprule
% \multirow{2}{*}{} & \multicolumn{4}{c}{Empathy}                      & Fluency \\ \cmidrule(l){2-6} 
%                   & EMOACC & IP   & EX   & \multicolumn{1}{c|}{ER}   & PPL     \\ \midrule
% k=2               & 0.24   & 0.08 & \textbf{0.57} & \multicolumn{1}{c|}{\textbf{1.10}} & \textbf{27.50}   \\
% k=3               & 0.25   & 0.09 & 0.48 & \multicolumn{1}{c|}{1.05} & 28.28   \\
% k=4               & \textbf{0.27}   & 0.09 & 0.40 & \multicolumn{1}{c|}{1.04} & 28.71   \\
% k=5               & 0.25   & \textbf{0.10} & 0.33 & \multicolumn{1}{c|}{1.00} & 28.49   \\
% k=6               & 0.25   & 0.08 & 0.32 & \multicolumn{1}{c|}{1.01} & 29.17   \\ \bottomrule
% \end{tabular}
% \label{tab:my-table4}
% \end{table}

\section{Evaluations on ChatGPT-Based Response Generation}
Then, we evaluate the responses generated by ChatGPT.
\subsection{Evaluation Models}
%\noindent\textbf{EmpGPT-3} \cite{lee2022does}: This approach proposes a method for selecting relevant in-context examples based on the situation and emotion as the prompt to improve GPT3's empathetic dialogue generation. We adopt the situation-based selection as a baseline, which computes the semantic similarity between the situation of the test and training data. Subsequently, selecting the top-k examples from the training data with the highest similarity score serve as in-context examples and prompt the GPT3 model.

%\noindent\textbf{EmpGPT-3+Causality}:

\noindent\textbf{ChatGPT}: The prompt given to ChatGPT includes only the chosen in-context raw examples $\mathcal S$ from the training set, along with the test sample.

\noindent\textbf{ChatGPT+Causality$_{user,sys}$}: The commonsense-based causality explanation prompt $\mathcal{M}_{prompt}^+$ is utilized to generate a response by ChatGPT, as illustrated in Algorithm \ref{alg:cap}.

\subsection{Evaluation Metrics}

\subsubsection{Automatic Metrics}
% In order to assess the quality of the GPT3/ChatGPT-based model, standard evaluation metrics as described in Section \ref{T5 automatic} are not appropriate. These metrics are based on evaluating the matching between the generated response and the ground truth, which is not applicable to GPT3/ChatGPT-based models as they do not undergo a traditional training process. Therefore, we adopt the following evaluations as the alternative metrics.% to assess the quality of the responses rather than their matching to a reference response.

% In addition to the evaluation metrics as described in Section \ref{T5 automatic}, we adopt the following evaluations to ChatGPT-based methods as they do not undergo a traditional training process.

\noindent\textbf{EMOACC}: Following \citet{welivita2020taxonomy,lee2022does}, we utilize the EMOACC \footnote{\url{https://github.com//passing2961/EmpGPT-3}} to measure the emotion accuracy of the generated responses, which is a fine-tuned BERT-base \cite{devlin2018bert} model on the EmpatheticDialogue dataset. 

\noindent\textbf{EMPTOME} \cite{sharma2020computational}: It consists of three empathy metrics: \textbf{Interpretations (IP)}, which represent expressions of acknowledgments or understanding of the interlocutor's emotion or situation. For example, a response like \textit{"I also worked hard for the math exam, which made me anxious,"} is considered a stronger interpretation than \textit{"I understand how you feel."} \textbf{Explorations (EX)}, which represent expressions of active interest in the interlocutor's situation. For instance, a statement like \textit{"Are you feeling terrified right now?"} exhibits stronger exploration compared to \textit{"What happened?"} \textbf{Emotional Reactions (ER)}, which represent expressions of explicit emotions. 
% a statement like \textit{"I feel excited for you"} with an explicitly labeled emotion is considered to have a stronger emotional reaction compared to \textit{"I am sorry to hear that."} 
% For instance, a statement with an explicitly labeled emotion is considered to have a stronger emotional reaction.
% such as warmth, compassion, and concern for the interlocutor's situation. 
They are computed by pre-trained empathy identification models.\footnote{\url{https://github.com/behavioral-data/Empathy-Mental-Health}} Specifically, RoBERTa \cite{liu2019roberta} models are separately fine-tuned for each metric by evaluating the generated response to the number of 0, 1, or 2, a higher value means stronger empathy.

\noindent\textbf{Coherence}: We leverage the BERTScore \cite{zhang2019bertscore} to quantify coherence by computing the semantic similarity between the generated response and the input context.
% thereby evaluating the coherence of the response with respect to the given context. %We adopt its matching precision, recall, and F1 score (PBERT, RBERT, FBERT). And we use the RoBERTa-Large \cite{liu2019roberta} version.

% \noindent\textbf{PPL}: Following \citet{pang2020towards, lee2022does}, we adopt the GPT2-XL to measure the fluency of the generated response through perplexity. 
% \noindent\textbf{Human A/B Test}: Comparing pairs of responses by different methods on the aspects of Coherence, Empathy, and Informativeness.

\subsubsection{Human A/B Test}
% In order to directly compare the performance of our proposed method with others, we conducted a human A/B test. For each test, two generated responses were presented to human evaluators - one generated by our T5+Casuality method and the other from one of the compared models: CEM, LEMPEx, or CARE. The results of previous human evaluations indicate that CEM and CARE were preferred over MoEL, MIME, and EmpDG, while LEMPEx outperformed MIME and EmpDG. Therefore, we only compared our method to CEM, LEMPEx, and CARE in this human evaluation.
% In order to directly compare the performance of our proposed method with others, we conducted a human A/B test. 

We also conducted A/B test to compare the performance of \textit{ChatGPT+Causality$_{user,sys}$} and \textit{ChatGPT}.
For each comparison, three crowd-workers are asked to choose the better one or select "Tie" based on three aspects: Empathy, Coherence, and Informativeness \cite{sabour2022cem}.
(1) \textbf{Empathy (Emp.)} measures whether the generated response understands the user's feelings and experiences. (2) \textbf{Coherence (Coh.)} measures whether the response is coherent/relevant in context. 
% and whether the topic is consistent with the context. 
(3) \textbf{Informativeness (Inf.)} evaluates whether the generated response conveys more information corresponding to the context. %For example, the response providing concrete solutions or suggestions is thought to be more informative. To be noted, the longer sentence doesn't mean more informative; the provided information should be reasonable and needed based on the user's utterance, rather the response is thought to be less informative.

\subsection{Results and Analysis}

\subsubsection{Number of In-context Examples}
We investigate the effect of the number of in-context examples using our proposed commonsense-based causality explanation prompt.
Table \ref{tab:my-table4} shows that setting $k$ to 4 results in the highest emotion accuracy, %while setting $k$ to 5 generates responses that better interpret the user's experiences and feelings, and setting $k$ to 6 leads to generating response closer to the ground truth. These improvements are attributed to ChatGPT's ability to learn more from the given examples. 
and setting $k$ to 2 yields better exploration and emotional reactions.
Therefore, we select $k$ values of 2 and 4 for the experiments. 
%Exploration represents the understanding of the seeker by exploring the feelings and experiences not stated in the post \cite{sharma2020computational}. Therefore, when given fewer examples, the model is encouraged to explore the topic in greater depth.

\begin{table}[ht]
\caption{Ablation study on the number of in-context examples $k$ in the prompt. 
% EMOACC, IP, EX, and ER represent emotion accuracy, interpretation, exploration, and emotion reaction, respectively.
}
\centering
\begin{tabular}{@{}lrrrrrr@{}}
\toprule
% \multirow{2}{*}{} & \multicolumn{4}{c}{Empathy}                      \\ \cmidrule(l){2-5} 
                  & EMOACC & IP   & EX   & \multicolumn{1}{r}{ER}      \\ \midrule
$k$=2               & 0.24   & 0.08 & \textbf{0.57} & \multicolumn{1}{r}{\textbf{1.10}} 
                % & 2.02 &1.25  
                \\
$k$=3               & 0.25   & 0.09 & 0.48 & \multicolumn{1}{r}{1.05} 
                % & 2.03 & 1.17   
                \\
$k$=4               & \textbf{0.27}   & 0.09 & 0.40 & \multicolumn{1}{r}{1.04} 
                % & 2.06 &1.30   
                \\
$k$=5               & 0.25   & \textbf{0.10} & 0.33 & \multicolumn{1}{r}{1.00} 
                % & 2.21 &1.44   
                \\
$k$=6               & 0.25   & 0.08 & 0.32 & \multicolumn{1}{r}{1.01} 
                % &\textbf{2.29} &\textbf{1.48}   
                \\ \bottomrule
\end{tabular}
\label{tab:my-table4}
\end{table}

\begin{table*}[t]
\caption{Evaluations on the effectiveness of causality$_{user,sys}$ when $k$ set to 2 and 4 with the single-turn setting for our ChatGPT-based methods.}
\centering
\scalebox{0.86}{
\begin{tabular}{@{}llrrrrrrr@{}}
\toprule
                     & \multirow{2}{*}{Method} & \multicolumn{4}{c}{Empathy}                            & \multicolumn{3}{c}{Coherence}                  \\ \cmidrule(l){3-9} 
                     &                         & EMOACC & IP     & EX     & \multicolumn{1}{r|}{ER}     & PBERT & RBERT & \multicolumn{1}{r}{FBERT}  \\ \midrule
\multirow{2}{*}{k=2} %& EmpGPT-3                   & \textbf{0.300}   & 0.032  & \textbf{1.144}  & \multicolumn{1}{c|}{0.944}  & 0.861  & 0.843  & \multicolumn{1}{c|}{0.852}  & 31.21   \\
                     %& EmpGPT3+Causality$_{user,sys}$        & 0.258 & 0.073 & 0.573 & \multicolumn{1}{c|}{\textbf{1.210}} & 0.855  & 0.854  & \multicolumn{1}{c|}{0.855}  & \textbf{23.99}   \\
                     & ChatGPT                 & 0.060   & 0.073  & 0.341  & \multicolumn{1}{r}{0.923}  & 0.877  & 0.872  & \multicolumn{1}{r}{0.875}  
                     % & 2.03 & 1.18 
                     \\
                     & ChatGPT+Causality$_{user,sys}$       & \textbf{0.280}   & \textbf{0.104}  & \textbf{0.768}  & \multicolumn{1}{c|}{\textbf{1.116}}  & \textbf{0.886}  & \textbf{0.878}  & \multicolumn{1}{r}{\textbf{0.882}}  
                     % & \textbf{2.15} & \textbf{1.30} 
                     \\ \midrule
\multirow{2}{*}{k=4} %& EmpGPT-3                   & 0.230   & 0.104  & 0.169  & \multicolumn{1}{c|}{0.956}  & 0.860  & 0.842  & \multicolumn{1}{c|}{0.851}  & 34.51   \\
                     & ChatGPT                 & 0.036  & 0.081  & 0.323  & \multicolumn{1}{r|}{0.867}  & 0.882  & \textbf{0.875}  & \multicolumn{1}{r}{0.879}  
                     % & \textbf{2.42}  & \textbf{1.45}
                     \\
                     & ChatGPT+Causality$_{user,sys}$       & \textbf{0.280}   & \textbf{0.120}  & \textbf{0.528}  & \multicolumn{1}{r|}{\textbf{1.076}}  & \textbf{0.888}  & 0.874  & \multicolumn{1}{r}{\textbf{0.881}}  
                     % & 2.10 & 1.28 
                     \\ \bottomrule
\end{tabular}}
\label{tab:my-table5}
\end{table*}

\begin{table*}[ht]
\caption{Evaluations on the effectiveness of causality$_{user,sys}$ when $k$ set to 2 and 4 with the multi-turn setting for our ChatGPT-based methods.}
\centering
\scalebox{0.86}{
\begin{tabular}{@{}llrrrrrrr@{}}
\toprule
                     & \multirow{2}{*}{Method} & \multicolumn{4}{c}{Empathy}                           & \multicolumn{3}{c}{Coherence}                 \\ \cmidrule(l){3-9} 
                     &                         & EMOACC & IP    & EX     & \multicolumn{1}{r|}{ER}     & PBERT & RBERT & \multicolumn{1}{c}{FBERT}   \\ \midrule
\multirow{2}{*}{k=2} %& EmpGPT-3                   & 0.235  & 0.051 & 0.217  & \multicolumn{1}{c|}{0.939}  & 0.875  & 0.851  & \multicolumn{1}{c|}{0.862}  & 35.08   \\
                     %& EmpGPT-3+Causality$_{user,sys}$          & 0.192 & 0.059 & 0.303 & \multicolumn{1}{c|}{1.092} & 0.875  & 0.855  & \multicolumn{1}{c|}{0.864}  & \textbf{23.67}   \\
                     & ChatGPT                 & 0.083  & \textbf{0.065} & 0.318  & \multicolumn{1}{r|}{0.917}  & 0.891  & 0.902  & \multicolumn{1}{r}{0.894}  
                     % & 1.51 & 0.79   
                     \\
                     & ChatGPT+Causality$_{user,sys}$       & \textbf{0.199}  & 0.058 & \textbf{0.397}  & \multicolumn{1}{r|}{\textbf{1.094}}  & \textbf{0.899}  & \textbf{0.907}  & \multicolumn{1}{r}{\textbf{0.901}}  
                     % & \textbf{1.91} & \textbf{1.21}   
                     \\ \midrule
\multirow{2}{*}{k=4} %& EmpGPT-3                   & 0.179  & \textbf{0.143} & 0.000    & \multicolumn{1}{c|}{0.710}   & 0.874  & 0.849  & \multicolumn{1}{c|}{0.861}  & 46.95   \\
                     & ChatGPT                 & 0.062  & \textbf{0.072} & \textbf{0.297}  & \multicolumn{1}{c|}{0.866}  & 0.896  & 0.904  & \multicolumn{1}{r}{0.898}  
                     % & 1.75 &0.94   
                     \\
                     & ChatGPT+Causality$_{user,sys}$       & \textbf{0.256}  & 0.065 & 0.282  & \multicolumn{1}{r|}{\textbf{1.007}}  & \textbf{0.902}  & \textbf{0.904}  & \multicolumn{1}{r}{\textbf{0.901}}  
                     % & \textbf{2.02} & \textbf{1.31}  
                     \\ \bottomrule
\end{tabular}}
\label{tab:my-table6}
\end{table*}

\begin{table}[ht]
\caption{Human A/B test when $k$ set to 2 and 4 with the single-turn setting for our ChatGPT-based methods. 
% "Coh.", "Emp.", and "Inf." represents coherence, empathy, and informativeness, respectively.
}
\scalebox{0.8}{
\begin{tabular}{lrrrrr}
\hline
Comparisons                                                                             & Aspects & Win  & Loss & Tie \\ \hline
\multirow{3}{*}{\begin{tabular}[c]{@{}l@{}}ChatGPT+Causality$_{user,sys}$ \\vs.  ChatGPT ($k$=2)\end{tabular}}   & Emp.    &  \textbf{50.7}   & 36.0     &  13.3\\
                                                                                               
                                                                                                & Coh.    & \textbf{42.7}    & 42.0     &  15.3 \\
                                                                                               & Inf.    &  \textbf{51.3}   & 37.3     &  11.3   \\ \hline
\multirow{3}{*}{\begin{tabular}[c]{@{}l@{}}ChatGPT+Causality$_{user,sys}$ \\vs. ChatGTP ($k$=4)\end{tabular}}     & Emp.    & \textbf{49.3}    &  32.7    &  18.0   \\
                                                                                               & Coh.    & 20.0    &  24.0    & \textbf{56.0}    \\
                                                                                               & Inf.    & \textbf{43.3}     & 40.7     & 16.0    \\ \hline
\end{tabular}}
\label{tab:my-table7}
\end{table}

\begin{table*}[t]
\caption{Automatic evaluation results of baselines and our T5-based method. Bold denotes the best score.}
\centering
\scalebox{0.83}{
\begin{tabular}{@{}llrrrrrrrrrr@{}}
\toprule
                                  & Methods                                                   & PPL $\downarrow$  & BLEU-2 & BLEU-3 & BLEU-4 & D1   & D2   & PBERT & RBERT & FBERT \\ \midrule
\multirow{6}{*}{Baselines}     & MOEL                                                     & 37.63  & 8.63   & 4.25   & 2.43   & 0.38 & 1.74  & 86.19   & 85.67   & 85.91    \\
                                  & MIME                                                     & 36.84  & 8.37   & 4.31   & 2.51   & 0.28 & 0.95  & 86.27   & 85.59   & 85.92    \\
                                  & EmpDG                                                     & 38.08   & 7.74   & 4.09   & 2.49   & 0.46 & 1.90  & 86.09   & 85.49   & 85.78    \\
                                  & CEM                                                      & 36.36  & 6.35   & 3.55   & 2.26   & 0.54 & 2.38  & 86.61   & 85.39   & 85.98    \\
                                  & LEMPEx                                                  & 30.42    & 2.1    & 0.8    & 0.35   & 1.02 & \textbf{10.81} & 83.60   & 83.09   & 83.34    \\
                                  % & CARE                                                    & 32.84   & 9.24   & 4.88   & 2.95   & 0.58 & 2.31  & 86.57   & 85.80   & 86.17    \\
                                  \midrule
\multirow{3}{*}{\begin{tabular}[c]{@{}l@{}}Ours\end{tabular}} & T5                                                        & 46.13    & 3.59   & 1.94   & 1.15   & 0.49 & 2.82  & 86.69   & 84.07   & 85.35    \\
                                  & T5+Causality$_{user}$                           & 15.26   & 4.84   & 1.97   & 0.89   & \textbf{1.08} & 10.75 & 90.16   & 89.48   & 89.80    \\
                                  & T5+Causality$_{user,sys}$ & \textbf{13.07}   & \textbf{10.53}  & \textbf{6.34}   & \textbf{4.06}   & 0.75 & 5.52  & \textbf{92.24}   & \textbf{90.76}   & \textbf{91.48}    \\ \bottomrule 
\end{tabular}}
\label{tab:my-table8}
\end{table*}

% \begin{table*}[ht]
% \caption{Automatic evaluation results of T5+Causality$_{user,sys}$ and ChatGPT+Causality$_{user,sys}$. D1 and D2 represent diversity-1 and diversity-2, respectively.}
% \scalebox{0.89}{
% \begin{tabular}{@{}lccccccccc@{}}
% \toprule
% % Model             & EMOACC & IP    & EX    & ER    & D1   & D2    & BLEU-2 & BLEU-3 & BLEU-4 \\ \midrule
% \multirow{2}{*}{Model} & \multicolumn{4}{c}{Empathy}    & \multicolumn{2}{c}{Diversity} & \multicolumn{3}{c}{BLEU} \\ \cmidrule(l){2-10} 
%                        & EMOACC & IP    & EX    & ER    & D1            & D2            & BLEU-2 & BLEU-3 & BLEU-4 \\ \midrule
% T5+Causality$_{user,sys}$      & 0.125  & \textbf{0.271} & 0.498 & 0.751 & 0.75 & 5.52  & \textbf{10.53}  & \textbf{6.34}   & \textbf{4.06}   \\
% ChatGPT+Causality$_{user,sys}$ & \textbf{0.235}  & 0.046 & \textbf{0.668} & \textbf{1.109} & \textbf{2.91} & \textbf{16.44} & 3.95   & 2.17   & 1.32   \\ 
% % Ground Truth & \textbf{0.190}  & 0.279 & \textbf{0.688} & \textbf{0.501} & \textbf{6.49} & \textbf{38.49} & -   & -   & -   \\
% \bottomrule
% \end{tabular}
% \label{tab:my-table16}
% }
% \end{table*}

\begin{table}[t]
\centering
\caption{Results of human A/B test for our T5-based model. }
\scalebox{0.9}{
\begin{tabular}{@{}lrrrr@{}}
\toprule
Comparisons             & Aspects         & Win & Loss & Tie \\ \midrule
\multirow{3}{*}{{\begin{tabular}[c]{@{}l@{}}T5+Causality$_{user,sys}$\\ vs. CEM\end{tabular}}}    & Emp.         & \textbf{42.0}    & 40.0     & 18.0\\
& Coh.       & \textbf{38.7}    & 33.3     & 28.0   
                         \\
                      & Inf. & 38.3    & \textbf{44.3}     & 17.3    \\ 
                  
                      \midrule
\multirow{3}{*}{{\begin{tabular}[c]{@{}l@{}}T5+Causality$_{user,sys}$\\ vs. LEMPEx\end{tabular}}}             & Emp.            & \textbf{53.0}    & 35.0     &  12.0   \\  & Coh.       & \textbf{39.0}    & 33.3     & 27.7    \\
                      
                      & Inf.           & \textbf{50.0}    &  38.0    & 12.0    \\ 
%                       \midrule
% \multirow{3}{*}{{\begin{tabular}[c]{@{}l@{}}ChatGPT+Causality$_{user,sys}$\\ vs. ChatGPT\end{tabular}}} & Coh.        &\textbf{42.7}    & 42.0     &  15.3     \\
%                       & Emp.         &\textbf{50.7}   & 36.0     &  13.3     \\
%                       & Inf.  &\textbf{51.3}   & 37.3     &  11.3     \\ 
\bottomrule
\end{tabular}}
\label{tab:my-table9}
\end{table}

\begin{table}[ht]
\caption{Evaluation results of the responses generated by our T5-based method and baselines. The closest to the ground truth is marked as bold.
% EX, IP, ER represents exploration, interpretation and emotion reaction, respectively.
}
\scalebox{0.82}{
\begin{tabular}{@{}lrrrr@{}}
\toprule
Methods      & EMOACC & IP    & EX    & ER    \\ \midrule
MoEL         & 0.103  & 0.184 & 0.209 & 1.166 \\
MIME         & 0.076  & 0.099 & 0.207 & 1.256 \\
EmpDG        & 0.091  & 0.150 & 0.169 & 1.270 \\
CEM          & 0.091  & 0.091 & 0.569 & 0.950 \\
LEMPEx       & 0.090  & 0.135 & 0.861 & \textbf{0.575} \\ \midrule
T5           & 0.049  & 0.110 & 0.408 & 1.299 \\
T5+Causality$_{user}$  & 0.093  & 0.172 & \textbf{0.685} & 0.784 \\
T5+Causality$_{user,sys}$ & \textbf{0.125}  & \textbf{0.271} & 0.498 & 0.751 \\ \midrule
Ground Truth & 0.190  & 0.279 & 0.688 & 0.501  \\ 
\bottomrule
\end{tabular}}
\label{tab:my-table15}
\end{table}

\begin{table}[ht]
\caption{Automatic evaluation results of T5+Causality$_{user,sys}$ and ChatGPT+Causality$_{user,sys}$ ($k$=2, with whole test set and both single and multi-turn settings). }
\scalebox{0.76}{
\begin{tabular}{@{}lrrr@{}}
\toprule
\multirow{2}{*}{Evaluations} & \multicolumn{1}{l}{} & T5+                                           & ChatGPT+                                      \\
                             & \multicolumn{1}{l}{} & \multicolumn{1}{l}{Causality$_{user,sys}$} & \multicolumn{1}{l}{Causality$_{user,sys}$} \\ \midrule
\multirow{4}{*}{Empathy}   & EMOACC & 0.125 & \textbf{0.235}   \\
                           & IP     & \textbf{0.271} & 0.046   \\
                           & EX     & 0.498 & \textbf{0.668}   \\
                           & ER     & 0.751 & \textbf{1.109}   \\ \midrule
\multirow{2}{*}{Diversity} & D1     & 0.75  & \textbf{2.91}    \\
                           & D2     & 5.52  & \textbf{16.44}   \\ \midrule
\multirow{3}{*}{BLEU}      & BLEU-2 & \textbf{10.53} & 3.95    \\
                           & BLEU-3 & \textbf{6.34}  & 2.17    \\
                           & BLEU-4 & \textbf{4.06}  & 1.32    \\ \bottomrule
\end{tabular}
\label{tab:my-table16}
}
\end{table}

\subsubsection{Experimental Results}
Table \ref{tab:my-table5} and Table \ref{tab:my-table6} present the results of 
% comparisons between 
\textit{ChatGPT} and \textit{ChatGPT+Causality$_{user,sys}$} with $k$ set to 2 and 4, under the single-turn and multi-turn settings, respectively. In the single-turn setting, a test sample consists of one utterance, while in the multi-turn setting, a test sample contains multiple turns. 
From the four comparisons, we observe that \textit{ChatGPT+Causality$_{user,sys}$} outperforms \textit{ChatGPT} in at least 5 out of 7 evaluation metrics. Notably, \textit{ChatGPT+Causality$_{user,sys}$} significantly outperforms \textit{ChatGPT} on \textit{EMOACC} and \textit{ER}, indicating that \textit{ChatGPT+Causality$_{user,sys}$} can generate responses with appropriate emotions. This can be attributed to the inclusion of inferred user emotions and reasoned system emotions, which provide appropriate affective information for generating empathetic responses. This improvement addresses the limitation of \textit{ChatGPT} on emotion recognition, as highlighted in \citet{zhao2023chatgpt}.
% which reports a performance gap of 3-18 percentage points compared to state-of-the-art fine-tuned models on emotion recognition tasks across four popular datasets. when reasoning from the system's perspective before responding. 

\textit{ChatGPT+Causality$_{user,sys}$} performs better when $k$ is set to 2 under the single-turn setting. Overall, the performance of \textit{ChatGPT+Causality$_{user,sys}$} is superior in the single-turn setting compared to the multi-turn setting. 
% This discrepancy can be attributed to the fact that COMET, which is employed for knowledge inferring when taking the in-context learning for causality prediction, is trained on event-based cause-effect graphs rather than context reasoning, making it less effective in predicting causality based on longer contexts. To solve the limitation of COMET will be put as our future work.
This discrepancy can be attributed to COMET, which is trained based on events, not context, making it less effective in predicting causality for long context. 
To solve the limitation of COMET will be placed on our future work.
% fact that COMET, which is used for knowledge inferring when constructing our causality explanation module, is trained based on events, not context, making it less effective in predicting causality for long context. To solve the limitation of COMET will be put as our future work.

The results of the human A/B test in Table \ref{tab:my-table7} show that \textit{ChatGPT+Causality$_{user, sys}$} is better than \textit{ChatGPT} on the aspects of \textit{Empathy} and \textit{Informativeness} because of the enriched knowledge by the commonsense-based causality explanations.

% \subsubsection{Case Studies}
% Table \ref{tab:my-table12} shows a case about the comparison between \textit{ChatGPT} and \textit{ChatGPT+Causality$_{user,sys}$}, and illustrates the impact of our proposed commonsense-based causality explanation. We can see that both the responses by \textit{ChatGPT} and \textit{ChatGPT+Causality$_{user,sys}$} show emotion reactions to the user's context. 
% % However, \textit{ChatGPT+Causality$_{user,sys}$} outperforms \textit{ChatGPT} by providing detailed suggestions that align with the user's desires based on reasoned intentions. As discussed in Section \ref{sec5.5.1}, COMET is not always reliable in its predictions. This sensitivity is evident in Table \ref{tab:my-table11}, where the inferred desires of the user mislead the reasoned intentions of the system.

\section{Experiments on T5-Based Response Generation}
Finally, we evaluate the responses generated by the T5-based model.
% We evaluate our approach on the EmpatheticDialogue dataset.
\subsection{Evaluation Metrics}
(1) Perplexity (PPL) \cite{vinyals2015neural} which measures the confidence of the generated response. (2)BLEU. (3) D1/D2 (Distinct-1/ Distinct-2) \cite{li2016diversity} which evaluates the diversity aspect. (4)BERTscore. (5) Human A/B Test.

\subsection{Evaluation Models}

\noindent\textbf{Affection-based Methods}: MoEL \cite{lin2019moel}; MIME \cite{majumder2020mime}; EmpDG \cite{li2020empdg}.

% \noindent\textbf{MoEL} \cite{lin2019moel}:
% This is an extension of Transformer, which softly combines multiple emotion-specific decoders to a meta decoder to generate an empathetic response.

% % \subsubsection{Human Rating}
% % To explore the fine-grained attributions of user's attributes (want and reaction by Comet) and sys's attributes (intention and reaction by ChatGPT) on the aspects of empathy, coherence, and informativeness, we randomly sample 100 dialogues and their corresponding generations from \textit{T5}, \textit{T5+Causality$_{user}$}, \textit{T5+Causality$_{user,sys}$} for comparison. Annotators are asked to evaluate the quality of the generated response based on three dimensions, and each metric is rated on a scale from 1 to 5.

% \noindent\textbf{MIME} \cite{majumder2020mime}:
% This method %assumes that empathetic responses often mimic the speaker's emotion and 
% integrates emotion grouping, emotion mimicry, and stochasticity into the emotion mixture for various empathetic responses.

% \noindent\textbf{EmpDG} \cite{li2020empdg}: %The model utilizes the emotion lexicon to detect nuanced emotions at the word-level, which are then integrated into its decoder inputs. 
% This model detects nuanced emotions and integrates them into the decoder. And it employs an emotional discriminator and a semantic discriminator to incorporate user feedback.

\noindent\textbf{COMET-based Method}: CEM \cite{sabour2022cem}, which employs commonsense knowledge, such as the user's reactions, intentions, desires, needs, and effects, to enhance its understanding of the interlocutor's situations and emotions.

% \noindent\textbf{CEM} \cite{sabour2022cem}: This model employs commonsense knowledge, such as the user's reactions, intentions, desires, needs, and effects, to enhance its understanding of the interlocutor's situations and emotions.

\noindent\textbf{T5-based Method}: LEMPEx \cite{majumder2022exemplars}, which adopts T5 as the encoder-decoder and utilizes a combination of exemplar-based retrieval, a response generator, and an empathy control module to generate empathetic responses.% that take into account the communication elements of the interlocutors.

% \noindent\textbf{CARE} \cite{wang2022care}: This model constructs the causality relationship of the conversation based on the Cause Effect Graph \cite{li2021guided} and utilizes the conditional graph prediction to achieve the reasoning. 

\noindent\textbf{T5} \cite{raffel2020exploring}: We utilize the T5 model as our base encoder-decoder architecture, integrating with the emotion classifier.
 We train it from scratch on the EmpatheticDialogue dataset. %which can be considered a fair comparison with other methods.
 
\noindent\textbf{T5+Causality$_{user}$}: The T5 model is extended with an additional T5 encoder for user's desires/reactions.

\noindent\textbf{T5+Causality$_{user,sys}$}: The T5 model is extended with two T5 encoders for the user's causality attributes (desires/reactions) and the system's causality attributes (intentions/reactions), respectively.

\subsection{Settings}
We trained T5-small \cite{raffel2020exploring} from scratch on the EmpatheticDialogues dataset. The learning rate is set to 0.00001, the batch size is set to 8, we utilize the top-$k$ search decoding strategy with $k$ set to 20, and sampling with the temperature set to 0.2, the max generation length set to 40.
%Adam \cite{kingma2014adam} is used as the optimizer.
% Causality$_{sys}$ is reasoned by ChatGPT with our commonsense-based causality explanation module.% prompt under $k$ set to 2 on the whole test set.

\subsection{Results and Analysis}
% In the A/B human test, we presented two responses to human evaluators - one generated by our T5+Casuality$_{user,sys}$ method and the other from one of the compared models: CEM and LEMPEx. 

Previous studies \cite{sabour2022cem,majumder2022exemplars} have shown that CEM and LEMPEx outperformed MoEL, MIME, and EmpDG. Therefore, we compared our method with CEM and LEMPEx in the human A/B test.
Automatic evaluation results shown in Table \ref{tab:my-table8} and human A/B test results shown in Table \ref{tab:my-table9} demonstrate the effectiveness of the proposed commonsense-based causality explanation (Causality$_{user,sys}$). 
The performance comparison presented in Table \ref{tab:my-table15} demonstrates the superiority of our method over the baselines in terms of emotion accuracy (EMOACC), interpretation (IP), and emotion reaction (EX) when compared to the ground truth.
% However, our method falls behind \textit{LEMPEx} in terms of exploration (EX). The concept of exploration refers to the model's ability to comprehend the seeker's emotions and experiences that may not be explicitly stated in the input post \cite{sharma2020computational}. And the exemplary-based retrieval mechanism introduced by \textit{LEMPEx} enabled them to enhance their model by leveraging related external information.

\subsection{Comparison between T5-based and ChatGPT-based Response Generation}
We conducted a performance comparison between the T5-based and ChatGPT-based response generation, as presented in Table \ref{tab:my-table16}. In terms of "Empathy," \textit{ChatGPT+Causality$_{user,sys}$} outperforms \textit{T5+Causality$_{user,sys}$} for EMOACC, EX, and ER, but performs worse for IP. Stronger interpretation (IP), which involves understanding and empathizing through shared experiences \cite{sharma2020computational}, is more frequently observed in the T5-based model, which was trained from the ground truth. In contrast, ChatGPT-based generation is not constrained by the ground truth and tends to respond from the perspective of a machine.

In terms of "Diversity" and "BLEU," it is evident that \textit{ChatGPT+Causality$_{user,sys}$} exhibits a larger diversity but results in a higher degree of mismatch with the ground truth (lower BLEU scores), indicating a potential need of balancing the response diversity and the accuracy in generating empathetic responses. 

Comparative case studies between T5-based and ChatGPT-based models with corresponding baselines can be seen in Appendix \ref{sec:case}.

% For instance, when given the input "I just moved to this neighborhood and some dumb criminals shot one of my neighbors and ran into the woods!", the response generated by \textit{T5+Causality${user,sys}$} is "I would have been so scared. What did you do? In comparison, the response generated by \textit{ChatGPT+Causality${user,sys}$} is "That's terrible. Did you call the police?" 

\section{Conclusions and Future Work}
We have proposed a commonsense-based causality explanation approach for diverse empathetic response generation that considers the system's intentions and reactions as well as the user's desires and reactions. Specifically, we enhance ChatGPT's ability to reason the system's intentions and reactions by integrating in-context learning with commonsense knowledge (desire, reaction, and intention). We have integrated the commonsense-based causality explanation with both ChatGPT and a trained T5 model. The experimental results demonstrate that our method outperforms other competitive methods on both automatic and human evaluations.

In the future, we will explore fine-grained approaches for causality explanation from the perspective of both the user and the system.

% \begin{table}[ht]
% \centering
% \caption{Human rating results of the ablation study for our method.}
% \begin{tabular}{@{}llll@{}}
% \toprule
% Model Variant             & Coh. & Emp. & Inf. \\ \midrule
% T5                        &      &      &      \\
% T5+Causality$_{user}$       &      &      &      \\
% T5+Causality$_{user,sys}$ &      &      &      \\ \bottomrule
% \end{tabular}
% \label{tab:my-table10}
% \end{table}

% \subsection{Appendices}

% Use \verb|\appendix| before any appendix section to switch the section numbering over to letters. See Appendix~\ref{sec:appendix} for an example.

\section*{Acknowledgements}
This work was supported by JST Moonshot R\&D Grant Number JPMJMS2011. This work was also supported by JST, the establishment of university fellowships towards the creation of science and technology innovation, Grant Number JPMJFS2123.

\bibliography{anthology,custom}
\bibliographystyle{acl_natbib}

\appendix

\section{Case Analysis on the COMET} \label{sec5.5.1}
We evaluate the effectiveness of COMET in inferring intents and reactions since ChatGPT's ability to reason them is sensitive to the given in-context examples. We assess 60 samples from the EmpatheticDialogue dataset based on two evaluation metrics: (1) Whether the inferred intents or reactions capture the context; (2) whether there are any conflicts among the generated intents or reactions. 
% The results are presented in Table \ref{tab:my-table2} and Table \ref{fig:my_label3} for intents and reactions, respectively.
We find that 51 out of 60 intent predictions and 46 out of 60 reaction predictions are acceptable. Table \ref{tab:my-table13} and \ref{tab:my-table14} show the example of reasoned intentions and reactions, respectively.

\begin{table}[ht]
\caption{Example intents inferred from COMET}
\scalebox{0.78}{
\begin{tabular}{@{}l@{}}
\toprule
An accepted example:                                                                                                           \\
sys: Did you suffer any injuries?                                                                                             \\
sys’s intents: to make sure they are ok; to know if you are ok. \\ \midrule
An unaccepted example that does not satisfy metric (1)                                                                        \\
sys: I understand that one, they are my favorite place to eat.                                                                \\
sys’s intents: to eat food; to eat good.                                               \\ \midrule
An unaccepted example that does not satisfy metric (2)                                                                        \\
sys: Jeez! It's so unfortunate... very sad really.                                                                         \\
sys’s intents: to be sad; to be happy. \\ \bottomrule
\end{tabular}}
\label{tab:my-table13}
\end{table}

\begin{table}[ht]
\caption{Example reactions referred by COMET}
\scalebox{0.85}{
\begin{tabular}{@{}l@{}}
\toprule
An accepted example                                                                                                 \\
sys: That's not good. Do you own a gun?                                                                             \\
sys’s reactions: scared; worried; nervous; fearful; angry                                                          \\ \midrule
An unaccepted example that does not satisfy metric (2)                                                              \\
\begin{tabular}[c]{@{}l@{}}sys: oh man. I'm all about discipline!\\ I don't like spoiled bratty kids.\end{tabular} \\
sys’s reactions: angry; good; happy; controlling; bad                                                              \\ \bottomrule
\end{tabular}}
\label{tab:my-table14}
\end{table}

\section{Introduction in the prompt for ChatGPT} \label{sec:introduction}
The introduction in the prompt for ChatGPT is shown in Table \ref{tab:my-table}, and the few-shot examples construction is in Table \ref{tab:my-table18}.

\section{Case Studies and Error Analysis}\label{sec:case}

Table \ref{tab:my-table12} shows a case about the comparison between \textit{ChatGPT} and \textit{ChatGPT+Causality$_{user,sys}$}, and illustrates the impact of our proposed commonsense-based causality explanation. We can see that both the responses by \textit{ChatGPT} and \textit{ChatGPT+Causality$_{user,sys}$} show emotion reactions to the user's context. However, \textit{ChatGPT+Causality$_{user,sys}$} outperforms \textit{ChatGPT} by providing detailed suggestions that align with the user's desires based on reasoned intentions. As discussed in Section \ref{sec5.5.1}, COMET is not always reliable in its predictions. This sensitivity is evident in Table \ref{tab:my-table11}, where the user's inferred desires mislead the reasoned intentions of the system.

Table \ref{tab:my-table17} further shows comparative case studies between T5-based and ChatGPT-based models with corresponding baselines.

\begin{table*}[t]
\caption{Introduction template to ChatGPT for causality reasoning and empathetic response generation.}
\label{tab:my-table}
\scalebox{0.86}{
\begin{tabular}{@{}l@{}}
\toprule
Introduction:                                                                                                                                         \\
\begin{tabular}[c]{@{}l@{}}Assuming that you are sys, who is a friend of the user. You are empathetic sometimes.\\ In this task, you are given the user's input and the information of "user wants to:" and "user reacts to:":\\ "user wants to:", which means what the user wants to do after the input;\\ "user reacts to:", which means how the user react to the input.\\ \\ After that, please reason about the following two parts:\\ "sys's intent:": which means what the sys wants to do after the input, or what's the intent of sys to respond to the input;\\ "sys reacts to:", which means how the sys reacts to the input.\\ \\ Then you respond (should be concise, no more than 30 words) to the input based on the information \\ of user's input, "user wants to:", "user reacts to:", "sys's intent:", "sys reacts to:".\\ \\ "sys:": which means the response of sys.\\ \\ Please generate the following three parts in the format below:\\ sys's intent:\\ sys reacts to:\\ sys:\end{tabular} \\ \bottomrule
\end{tabular}}
\end{table*}

\begin{table*}[]
\caption{Few-shot examples (top-$2$ examples). %"user wants," "user reacts to," "sys 's intent" and "sys reacts to" are the corresponding example causality for each example. And sys is the abbreviation of system.
}
\scalebox{0.84}{
\begin{tabular}{@{}lll@{}}
\toprule
Test input                & \multicolumn{2}{l}{user: I'm so excited because I'm finally going to visit my parents next month!  I didn't see them for 3 years.}                                                                                                                                   \\ \midrule
\multirow{6}{*}{Few-shot1} & context1                                                                     & \begin{tabular}[c]{@{}l@{}}user1: Someone is visiting me soon and I can't wait!\\ sys1: Who is it? \\ user1: My mom, she is amazing.\end{tabular}                                     \\ \cmidrule(l){3-3} 
                          & \multirow{4}{*}{\begin{tabular}[c]{@{}l@{}}example\\ causality\end{tabular}} & \textless{}xWant\textgreater{}$_{user1}$: to have a good time. to talk to their mom. to have fun with Mom.                                                           \\
                          &                                                                              & \textless{}xReact\textgreater{}$_{user1}$: excited. happy. satisfied. good. loved.                                                                                                         \\
                          &                                                                              & \textless{}xIntent\textgreater{}$_{sys}$: to be with her. to be loved. to be nice. happy.                                                                                                 \\
                          &                                                                              & \textless{}xReact\textgreater{}$_{sys}$: happy. excited. proud. good. loving.                                                                                                            \\ \cmidrule(l){3-3} 
                          & response1                                                                    & sys1: I bet she is! I am so glad you get to see her. Mom's are awesome!                                                                                                               \\ \midrule
\multirow{6}{*}{Few-shot2} & context2                                                                     & \begin{tabular}[c]{@{}l@{}}user2: My family is coming to visit!\\ sys2: Awesome.  When are they coming and for how long?\\ user2: They are coming next year from Africa!\end{tabular} \\ \cmidrule(l){3-3} 
                          & \multirow{4}{*}{\begin{tabular}[c]{@{}l@{}}example\\ causality\end{tabular}} & \textless{}xWant\textgreater{}$_{user2}$: to have a good time. to go to the airport. to have fun with the family.                                                   \\
                          &                                                                              & \textless{}xReact\textgreater{}$_{user2}$: happy. excited. happy. excited. loved.                                                                                                          \\
                          &                                                                              & \textless{}xIntent\textgreater{}$_{sys2}$: to see the sights. to be with family. to be with them. to have fun.                                                                             \\
                          &                                                                              & \textless{}xReact\textgreater{}$_{sys2}$: happy. excited. satisfied. tired. relieved.                                                                                                      \\\cmidrule(l){3-3} 
                          & response2                                                                    & sys2: That's a long trip.  I hope they have a good time.                                                                                                                              \\ \bottomrule
\end{tabular}}
\label{tab:my-table18}
\end{table*}

\begin{table*}[t]
\caption{Case studies of the proposed method and baseline. Blue texts represent cognition, and orange texts show affection. Desires$_{user}$ and Reactions$_{user}$ are inferred by COMET, while Intentions$_{sys}$ and Reactions$_{sys}$ are reasoned by ChatGPT+Causality$_{user,sys}$. "sys" is the abbreviation of system.}
\scalebox{0.773}{
\begin{tabular}{@{}ll@{}}
\toprule
Context                    & user: I'm so scared of bugs! i found one in my hair yesterday and almost died.                                                                        \\ \midrule
ChatGPT                    & sys: That's so scary! Did you manage to get it out?                                                                                                  \\ \midrule
Desires$_{user}$  & \begin{tabular}[c]{@{}l@{}}user wants: \textcolor{cyan}{to get rid of bugs}; to go to the doctor; to run away.\end{tabular} \\
Reactions$_{user}$ & user reacts to: scared; frightened; terrified; afraid; nervous.                                                                                       \\
Intentions$_{sys}$                           & \begin{tabular}[c]{@{}l@{}}sys's intent: to comfort; to help; \textcolor{cyan}{to offer advice}; to be supportive; \textcolor{cyan}{to avoid bugs.}\end{tabular}                        \\
Reactions$_{sys}$      & sys reacts to: empathetic. \textcolor{orange}{caring}. helpful. \textcolor{orange}{understanding}. supportive.                                                                               \\
\begin{tabular}[c]{@{}l@{}}ChatGPT+Causality$_{user,sys}$ \end{tabular} 
         & \begin{tabular}[c]{@{}l@{}}sys: \textcolor{orange}{I know how you feel, bugs are the worst.} \textcolor{cyan}{Have you tried using bug spray or wearing a hat outside?}\end{tabular}    \\ \bottomrule
\end{tabular}}
\label{tab:my-table12}
\end{table*}

\begin{table*}[ht]

\caption{Error analysis of COMET knowledge's negative impact on the causality explanation. }
\scalebox{0.96}{
\begin{tabular}{@{}ll@{}}
\toprule
Context                     & user: I’m so grateful for my family due to an incident with my friend.                                                                                     \\ \midrule
ChatGPT                     & sys: What happened with your friend?                                                                                                                      \\ \midrule
Desires$_{user}$   & \begin{tabular}[c]{@{}l@{}}user wants: \textcolor{cyan}{to make amends to their friend}; to make amends for the incident. \end{tabular} \\
Intentions$_{sys}$  & \begin{tabular}[c]{@{}l@{}}sys's intent: \textcolor{cyan}{to help}; to support; to be a good friend; to be there for user.\end{tabular}                                    \\
\begin{tabular}[c]{@{}l@{}}ChatGPT+Causality$_{user,sys}$ \end{tabular}          & sys: I'm here for you if you \textcolor{cyan}{need any help making amends with your friend}.                                                                                \\ 

% Ground Truth                & Oh really? What happened with your family and friend then?                                                                                                 \\
\bottomrule
\end{tabular}}
\label{tab:my-table11}
\end{table*}

\begin{table*}[]
\caption{Comparative case studies between T5-based and ChatGPT-based models with corresponding baselines. Meanwhile, Causality$_{user,sys}$ includes both Causality$_{user}$ and Causality$_{sys}$.}
\scalebox{0.81}{
\begin{tabular}{@{}llll@{}}
\toprule
                            & Context                                & \multicolumn{2}{l}{\begin{tabular}[c]{@{}l@{}}user: I'm so sad because I've read an article about a newborn girl who died \\ because her parents didn't believe in medication and doctors.\end{tabular}} \\ \midrule
\multirow{5}{*}{Baselines} & MOEL                                   & \multicolumn{2}{l}{sys: I am so sorry to hear that. I hope you are able to find her.}                                                                              \\
                           & MIME                                   & \multicolumn{2}{l}{sys: That is sad. I am sorry to hear that.}                                                                                                     \\
                           & EmpDG                                  & \multicolumn{2}{l}{sys: I am sorry to hear that. I hope you find out of it.}                                                                                       \\
                           & CEM                                    & \multicolumn{2}{l}{sys: Oh no, I am so sorry to hear that.}                                                                                                   \\
                           & LEMPEx                                 & \multicolumn{2}{l}{sys: I'm sorry to hear that. I'm so sorry to hear that.}                                                                                  \\ \midrule
\multirow{7}{*}{Ours}      & T5                                     &                                                      & sys: Oh no! I'm so sorry to hear that. What happened?                                                  \\ \cmidrule(l){3-4} 
                           & \multirow{3}{*}{T5+Causality$_{user}$}      & \multirow{2}{*}{Causality$_{user}$}                      & user wants: to tell their friends; to cry; to find out what happened.                    \\
                           &                                        &                                                      & user reacts to: sad; sorrowful; depressed; upset.                                                              \\
                           &                                        & Response                                             & sys: Oh, that's terrible. I am so sorry for your loss.                                                 \\ \cmidrule(l){3-4} 
                           & \multirow{3}{*}{T5+Causality$_{user,sys}$} & \multirow{2}{*}{Causality$_{sys}$}                       & sys's intent: to comfort spk1; to show empathy; to be supportive.                                           \\
                           &                                        &                                                      & sys reacts to: sad; upset; empathetic; sympathetic; concerned.                                                \\
                           &                                        & Response                                             & sys: I know what you mean, I would have been so mad!                                                   \\ \midrule
Baseline         & ChatGPT                                & \multicolumn{2}{l}{\begin{tabular}[c]{@{}l@{}}sys: That's heartbreaking. \\ It's important to trust medical professionals and seek help when needed.\end{tabular}}                                            \\
Ours                       & ChatGPT+Causality$_{user,sys}$                       & \multicolumn{2}{l}{\begin{tabular}[c]{@{}l@{}}sys: That's really heartbreaking. \\ I'm here for you if you need to talk or just need someone to listen.\end{tabular}}                                         \\ \bottomrule  

\toprule
                           & Context                                & \multicolumn{2}{l}{user: I think that the girl of my dreams likes somebody else. I feel very sad about it.}                                                                                                           \\ \midrule
\multirow{5}{*}{Baselines} & MOEL                                   & \multicolumn{2}{l}{I am sorry to hear that. I hope you can find a new one.}                                                                                                                                           \\
                           & MIME                                   & \multicolumn{2}{l}{I am sorry to hear that. I hope you can find a good relationship with her.}                                                                                                                        \\
                           & EmpDG                                  & \multicolumn{2}{l}{I am sorry to hear that.}                                                                                                                                                                          \\
                           & CEM                                    & \multicolumn{2}{l}{I am sorry to hear that.}                                                                                                                                                                         \\
                           & LEMPEx                                 & \multicolumn{2}{l}{I understand that. it's really stressful.}                                                                                                                                                      \\ \midrule
\multirow{7}{*}{Ours}      & T5                                     &                                                                                & I'm sorry to hear that. I hope you can find a new job.                                                                               \\ \cmidrule(l){3-4} 
                           & \multirow{3}{*}{T5+Causality$_{user}$}      & \multirow{2}{*}{Causality$_{user}$}                                                & user wants: to find a new girlfriend; to ask her out.                                                                                      \\
                           &                                        &                                                                                & user reacts to: sad; upset; unhappy; disappointed; angry.                                                                                     \\
                           &                                        & Response                                                                       & sys: what did he buy you?                                                                                                           \\ \cmidrule(l){3-4} 
                           & \multirow{3}{*}{T5+Causality$_{user,sys}$} & \multirow{2}{*}{Causality$_{sys}$}                                                 & sys's intent: to comfort; to encourage; to help find a new love.                                                         \\
                           &                                        &                                                                                & sys reacts to: sympathetic; caring; supportive; encouraging.                                                                      \\
                           &                                        & Response                                                                       &   \begin{tabular}[c]{@{}l@{}}sys: I know how you feel. \\I have been there before and it is hard to find a new one.   \end{tabular}                                              \\ \midrule
Baseline         & ChatGPT                                & \multicolumn{2}{l}{\begin{tabular}[c]{@{}l@{}}sys: I'm sorry to hear that.\\ It's tough when we have feelings for someone who doesn't feel the same way. \end{tabular}} \\
Ours                       & ChatGPT+Causality$_{user,sys}$                     & \multicolumn{2}{l}{\begin{tabular}[c]{@{}l@{}}sys: I'm sorry to hear that.\\ Maybe it's time to move on and find someone who appreciates you.\end{tabular}}                                                                                                    \\ \bottomrule

\end{tabular}}
\label{tab:my-table17}
\end{table*}

\end{document}